\pdfoutput=1

\documentclass[11pt]{article}

\usepackage[table]{xcolor} 
\usepackage[final]{acl}

\usepackage{times}
\usepackage{latexsym}

\usepackage[T1]{fontenc}

\usepackage[utf8]{inputenc}

\usepackage{microtype}

\usepackage{inconsolata}
\usepackage{bbding}
\usepackage{pifont}
\usepackage{xspace}
\usepackage{amsmath}
\newcommand{\our}{\textsc{AVerImaTeC}\xspace}
\newcommand{\prev}{\textsc{AVeriTeC}\xspace}
\usepackage{graphicx}
\usepackage{booktabs}

\usepackage{amsfonts}

\title{The Automatic Verification of Image-Text Claims (AVerImaTeC)
\\ Shared Task}

\author{
Rui Cao\textsuperscript{1}, Zhenyun Deng\textsuperscript{1}, Yulong Chen\textsuperscript{1}, 
\textbf{Michael Schlichtkrull\textsuperscript{1,2}, Andreas Vlachos\textsuperscript{1}}
\\
\textsuperscript{1}University of Cambridge, \\
\textsuperscript{2}Queen Mary University of London\\
  \texttt{
  \{rc990,zd302,yc632,av308\}@cam.ac.uk \
  m.schlichtkrull@qmul.ac.uk
  }\\
}

\begin{document}
\maketitle
\begin{abstract}
The Automatic Verification of Image-Text Claims (\our) shared task 
aims to advance system development for retrieving evidence and verifying real-world image-text claims.
Participants were allowed to either employ external knowledge sources, such as web search engines, or leverage the curated knowledge store provided by the organizers.
System performance was evaluated using the \our score, defined as a conditional verdict accuracy in which a verdict is considered correct only when the associated evidence score exceeds a predefined threshold.
The shared task attracted 14 submissions during the development phase and 6 submissions during the testing phase. 
All participating systems in the testing phase outperformed the baseline provided.
The winning team, HUMANE, achieved an \our score of 0.5455.
This paper provides a detailed description of the shared task, presents the complete evaluation results, and discusses key insights and lessons learned.
\end{abstract}

\section{Introduction}
Automated fact-checking (AFC) aims to develop effective systems for curbing the spread of misinformation at scale.
To support research in this area, several benchmark datasets have been proposed~\cite{DBLP:conf/naacl/ThorneVCM18,DBLP:conf/nips/AlyGST00CM21,DBLP:conf/nips/SchlichtkrullG023,DBLP:conf/emnlp/AlhindiPM18,DBLP:conf/sigir/YaoS0CH23,DBLP:conf/naacl/ChenKSDC24}, with the goal of advancing the effectiveness and interpretability of AFC systems.
However, existing AFC benchmarks focus almost exclusively on textual claims, overlooking the fact that online misinformation is increasingly media-heavy. 
Prior studies show that media can enhance perceived credibility~\cite{Newman2012NonprobativeP} and increase information exposure~\cite{doi:10.1177/0022243719881113}.
Recent evidence further suggests that approximately 80\% of online claims are multimodal, involving both text and media~\cite{DBLP:journals/corr/abs-2405-11697}, with images being the most prevalent modality.

While several AFC datasets target image-text claims, the majority are synthetic, constructed by manually manipulating either the textual or the visual modality of image-text pairs~\cite{DBLP:conf/emnlp/LuoDR21,DBLP:journals/ijmir/PapadopoulosKPP24,DBLP:conf/cvpr/JiaHZJCL23}. 
Owing to the distributional shift between synthetic data and naturally occurring content, model performance on these benchmarks may fail to faithfully reflect their effectiveness on real-world claims~\cite{DBLP:journals/corr/abs-2407-13488}. 
Only a limited number of benchmarks are based on real-world claims verified through fact-checking articles.
However, claims from these datasets often omit critical information required for verification, such as unresolved references or missing contextual details~\cite{DBLP:conf/emnlp/OusidhoumY022,DBLP:conf/nips/SchlichtkrullG023}. 
Furthermore, both synthetic and real-world benchmarks largely lack explicit evidence annotations, making it difficult to assess models’ reasoning processes.

\begin{figure*}[t]
     \centering
     \includegraphics[scale=0.45]{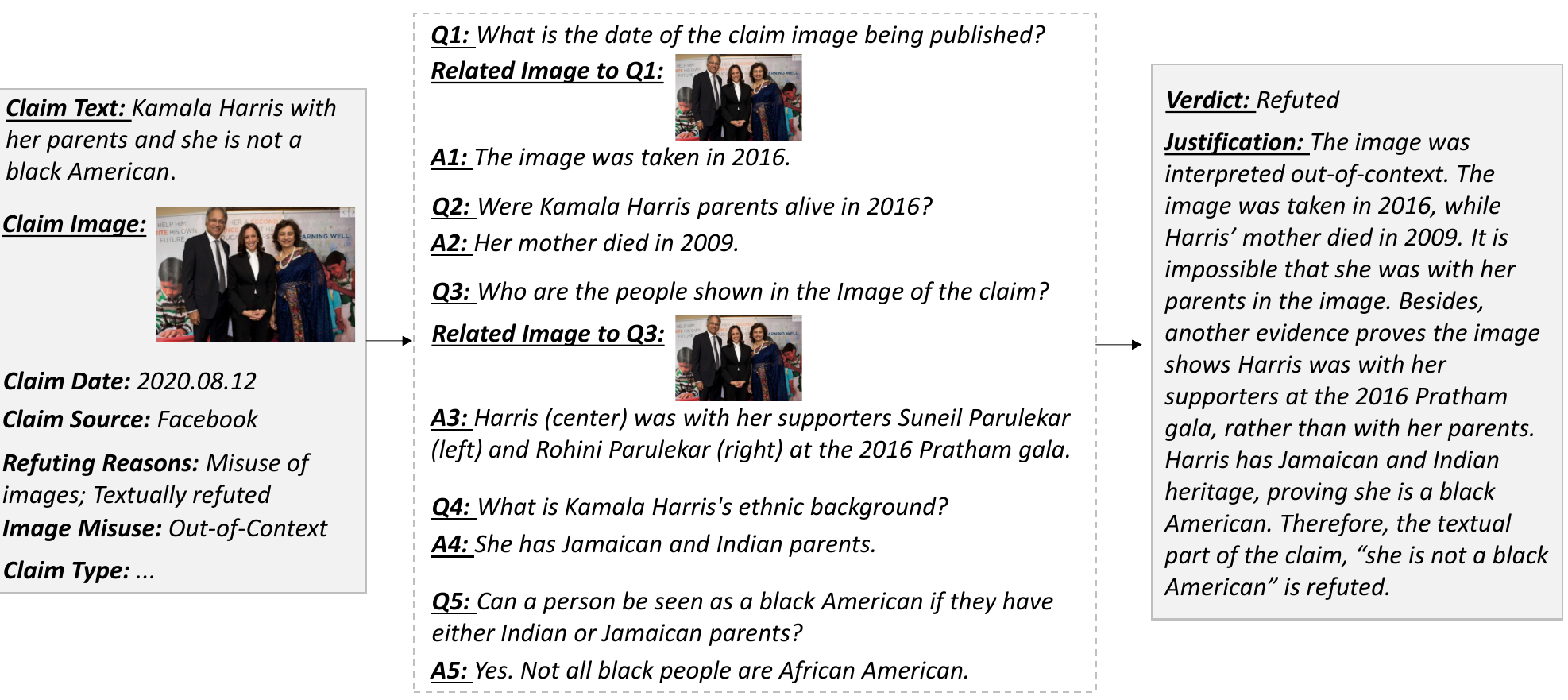}
     \caption{\textbf{An annotated claim from \our}. Given an image-text claim and its associated metadata, participating systems are required to first retrieve appropriate evidence, and subsequently predict a verdict accompanied by a textual justification grounded in the retrieved evidence.}
     \label{fig:claim-exp}
 \end{figure*}

To address the aforementioned limitations, the \our~\cite{DBLP:journals/corr/abs-2505-17978} dataset was introduced, and is the foundation for this year's FEVER shared task.
It comprises contextually independent image-text claims manually extracted from real-world fact-checking articles.
For each claim, the verification process is explicitly decomposed into a sequence of question-answer (QA) pairs, each supported by evidence retrieved from the web.
To mitigate temporal leakage identified in prior works~\cite{DBLP:conf/emnlp/OusidhoumY022,DBLP:conf/nips/SchlichtkrullG023}, 
All evidence associated with a claim is constrained to be published before the claim date. 
In addition, each claim is annotated with rich metadata, a verdict grounded in the retrieved evidence, and a textual justification explaining how the final verdict is reached. An example of an annotated claim is shown in Figure~\ref{fig:claim-exp}. The resulting dataset contains 1,297 claims.

The baseline published together with the dataset by \citet{DBLP:journals/corr/abs-2505-17978} uses a web search API to retrieve evidence.
Considering the cost associated with search APIs, we additionally release a curated knowledge store for the shared task.
For each claim, the knowledge store contains claim-relevant evidence sufficient for verification, along with adversarial and irrelevant evidence to simulate the noise and diversity of evidence retrieval from the open web.
We further develop an updated version of the shared-task baseline that is fully compatible with this knowledge store.
This design aims to alleviate the financial burden of web search APIs incurred during evidence retrieval. 
Participants are allowed to
1) rely solely on the provided knowledge store, 
2) retrieve evidence independently via external search engines at their own expense, 
or 3) combine both sources of evidence.
The dataset and baseline are released under a CC-BY-NC-4.0 license\footnote{https://fever.ai/dataset/averimatec.html}.

The shared task attracted 14 submissions in the development phase and 6 submissions in the testing phase. 
The primary evaluation criterion is conditional verdict prediction accuracy, which we refer to as the \our score, in which a system’s verdict prediction can only be counted as correct only if its associated evidence score exceeds a predefined threshold. 
In the testing phase, all submitted systems surpassed the provided baseline, demonstrating consistent advances in system design. The top-performing team, HUMANE, achieved an \our score of 0.5455, compared to 0.1136 for the baseline.

This paper first describes the shared task (Section~\ref{sec:task}), including the dataset, the provided knowledge store, the baseline system, and the evaluation protocol based on the \our score used for the leaderboard.
We then present an overview of the submitted systems in Section~\ref{sec:results}, analyzing their methodologies and performance, and highlighting key insights drawn from the submissions.
Finally, we reflect on the shared task and distill lessons for future research on real-world image-text claim verification (Section~\ref{sec:lessons}).

\begin{table*}[t]
\centering
\small
  \begin{tabular}{l|ccc}
    \toprule
    \textbf{Split} &\textbf{Train} &\textbf{Dev} &\textbf{Test} \\
    \midrule
    \# Claims & 793&152 &352 \\
    \# Images / Claim  & 1.49&1.38 & 1.38\\
    \# QA Pairs / Claim & 2.86& 2.84& 3.11\\
    Reannotated (\%) &15.0 &15.8 &9.4 \\
    End Date &31-05-2023 &31-07-2023 &21-03-2025 \\
    Labels (S/R/C/N) (\%) & 1.6/95.3/0.8/2.3& 2.6/92.8/0.7/3.9& 13.9/78.1/2.0/6.0\\
    Types (EP/MA/Cs/Nm) &85.4/21.1/5.7/3.2 & 91.4/14.5/0.7/3.9&93.5/30.1/3.4/1.4 \\
    Strategies (RIS/Ct/WE/IA/SD) & 50.2/30.0/87.5/20.7/26.4&57.9/24.3/89.5/24.3/20.4
    &67.3/16.8/84.7/21.3/26.4 \\
    \bottomrule
\end{tabular}
\caption{\textbf{Data statistics for dataset splits.} End date refers to the latest publication date of claims included in each split. The start date of each \textit{dev} and \textit{test} split corresponds to the end date of the preceding split. Claim label distributions are reported across four categories: \textit{supported} (S), \textit{refuted} (R), \textit{conflicting/cherry-picking} (C), and \textit{not enough evidence} (N). Claim type distributions (\%) are reported over Event/Property (EP), Media Analysis (MA), Causal (Cs) and Numerical (Nm) claims.
Fact-checking strategy distributions (\%) are reported over Reverse Image Search (RIS), Consultation (Ct), Written Evidence (WE), Image Analysis (IA), and Media Source Discovery (MSD). Low-frequency claim types and strategies are omitted. For example, satirical source identification accounts for only 1.9\% of training claims.
}
  \label{tab:data-statistics}
\end{table*}
\section{Task Description}
\label{sec:task}
In the \our shared task, participants are provided with image-text claims along with associated metadata (e.g., claim date and claim location).
The goal of the task is to encourage system designs that not only predict correct verdicts but also retrieve and present appropriate evidence.

For each claim, participants are required to submit retrieved evidence. Each piece of evidence must be accompanied by a URL pointing to the external source from which it was obtained. Based on the retrieved evidence, participants must predict a verdict selected from \textit{supported}, \textit{refuted}, \textit{not enough evidence} and \textit{conflicting/cherry-picking}.
In addition, systems should provide a textual justification explaining how the predicted verdict is derived from the retrieved evidence.

In the gold annotations, evidence was originally retrieved by formulating information-seeking questions. For consistency, participants are asked to submit such questions. 
However, the shared task does not restrict evidence retrieval to this paradigm, and the quality of the submitted questions is used for reference only. 
Instead, participants are required to submit evidence statements directly, which may be derived from QA pairs. 
Accordingly, the gold QA annotations are converted into evidence statements to form the ground-truth (GT) evidence set.
To accommodate the multimodal nature of image-text claim verification, evidence statements may themselves be multimodal. Each evidence statement is therefore separated into a textual component and an image component. For evidence involving images, images should be represented in the textual component using special image tokens (e.g., \texttt{[IMG\_1]}), while the corresponding images are provided in the image component encoded in base64 format.

\subsection{Dataset}
\label{sec:task-dataset}
In the shared task, participants are required to use the publicly available \our dataset for training and validation. The training and development splits are released to support system design and tuning, while the test set is kept hidden to ensure a fair evaluation. Notably, the test data consists of more recent claims, which helps mitigate potential data leakage arising from model pre-training.
Table~\ref{tab:data-statistics} presents the data statistics of the \our dataset.

Similar to \prev, the dataset used in the previous shared task on textual claim verification~\cite{DBLP:journals/corr/abs-2410-23850,akhtar-etal-2025-2nd}, event/property claims remain the most dominant claim type, and written evidence continues to be the most commonly used form of verification.
However, \our exhibits a notably different distribution of claim types, with \textit{Media Analysis} as a prevalent category, which appeared only infrequently in \prev. In addition, fact-checking strategies that explicitly focus on the visual component of claims, such as reverse image search, image analysis, and media source discovery, are used substantially more often. These statistics underscore the inherently multimodal nature of image-text claim verification and highlight the critical importance of rigorously verifying the image component of online claims.

\subsection{Knowledge Store}
\label{sec:task-ks}
As mentioned in the introduction, 
We released a dedicated knowledge store alongside the shared task to facilitate participation.
This knowledge store is designed to simulate an open-web retrieval environment while eliminating the need for costly API usage.
It contains a curated collection of claim-specific evidence drawn from the open web, encompassing both textual and visual materials, which aligns with the inherently multimodal nature of the verification process:

\noindent\textbf{Textual Evidence: }
Textual evidence in the knowledge store is sourced from two categories of documents: those relevant for verifying the textual component of a claim and those supporting verification of the image component.
For evidence targeting textual claim verification, we follow the knowledge store construction protocol adopted in prior shared tasks~\cite{DBLP:journals/corr/abs-2410-23850,akhtar-etal-2025-2nd}. Specifically, we first generate a diverse set of search queries using Gemini-2.5-Flash, conditioned on the claim content and the GT annotations of the textual question-answer pairs.
To improve robustness and reflect realistic retrieval challenges, we additionally construct adversarial queries by perturbing key entities, dates, and events in the claim. These adversarial queries are designed to retrieve plausible yet irrelevant documents, thereby increasing the difficulty and diversity of the evidence pool.
Details of the query construction process are provided in Appendix~\ref{sec:app-ks}.

Using the constructed queries, we employ the Google Search API\footnote{https://developers.google.com/custom-search/v1/overview} to retrieve URLs from the first page of results. Temporal constraints are applied during retrieval to ensure that returned documents were published before the claim date, thereby preventing temporal leakage.
Evidence sourced from fact-checking websites is excluded as well to avoid label leakage. 
In addition, human-annotated ground-truth evidence URLs are explicitly included. All collected URLs are subsequently deduplicated and randomly shuffled.

Retrieving image-related evidence is also essential for verifying the image component of image-text claims, as described in Section~\ref{sec:task-dataset}. To this end, we also collect textual documents associated with claim images during the knowledge store construction process.
Motivated by the widespread use of \textit{reverse image search} (RIS) in professional fact-checking workflows, we leverage the RIS functionality provided by Google Cloud Vision\footnote{https://docs.cloud.google.com/vision/docs}. 
Claim images are used as inputs to the RIS system, which returns URLs of web pages containing the same or visually similar images.
To avoid temporal leakage, we further filter out pages published prior to the claim date by extracting publication times using the \textit{html.find\_date} Python package.

Based on the collected URLs, we scraped the textual content with the package \textit{trafilature}~\cite{DBLP:conf/acl/Barbaresi21}. For URLs containing PDFs, we exploited the package, PyMuPDF, for textual content extraction.
\begin{table*}[ht]
    \centering
    \scalebox{0.75}{
\addtolength{\tabcolsep}{0pt}
    \begin{tabular}{l|cc|cc|c}
    \toprule
    
    \textbf{Split}& \multicolumn{4}{c|}{\textbf{Textual}} & \textbf{Image} \\
    \midrule
    & \multicolumn{2}{c|}{GS} & \multicolumn{2}{c|}{RIS} &GS\\
    & \# URLs & \# Words & \# URLs & \# Words &  \# Images \\
    
    \midrule
    Train &860,517/519,050 &2,909,977,889 &13,481/11,332 &6,866,742 &81,817\\
    Dev. & 163,684/99,452&573,148,913 &2,371/1,953 &1,142,074 & 49,617 \\
    Test &786,022/503,006 &2,622,659,482 &9,717/7,814 &4,387,002 & 116,860 \\
    \bottomrule
    \end{tabular}
    }
    \caption{\textbf{Statistics of textual evidence and image evidence in the provided knowledge store.}  For each URL entry, the first number denotes the total number of collected URLs, while the second indicates the number of URLs from which valid textual content was successfully scraped. GS denotes Google Search using textual queries, and RIS refers to reverse image search with images as input.}
\label{tab:stat_kb}
\end{table*}

\noindent\textbf{Image Evidence: }
In addition to textual documents, evidence in the knowledge store also includes images. 
To collect image-based evidence, we employ the Google Search API restricted to direct image URLs.
Specifically, we reuse the search queries generated during textual evidence collection and submit them to the API to retrieve image resource URLs.
Temporal constraints are applied as well.
Based on the retrieved URLs, we download the corresponding images. Considering both the storage cost associated with large image files and the relatively low frequency (only $1.6\%$ of human annotated answers are images) with which images are used as direct evidence, we cap the download to images from the top 100 retrieved URLs per query for the knowledge store of the training split. This design balances coverage with practical resource constraints.

\noindent\textbf{Knowledge Store Summary: }
We provide the statistics of the released knowledge store in Table~\ref{tab:stat_kb}. By providing a pre-collected repository of both textual and visual evidence, the knowledge store alleviates participants’ reliance on commercial search APIs, reduces financial barriers, and improves the reproducibility of the shared task and evaluated systems.
Beyond including URLs corresponding to human-annotated GT evidence, we further enrich the evidence pool by retrieving potentially relevant evidence
using query variants derived from the golden question-answer pairs.
This design introduces alternative evidence paths beyond the annotated gold evidence, offering participants greater flexibility in evidence selection and retrieval strategy design.

\subsection{Baseline}
The baseline system largely follows the design proposed in the original \our paper~\cite{DBLP:journals/corr/abs-2505-17978}, with targeted modifications to ensure compatibility with the knowledge store described in Section~\ref{sec:task-ks}. 
These updates are intended to lower the implementation burden for participants and facilitate faster onboarding to the shared task.

The baseline adopts a pipeline consisting of four components: a \textit{question generator}, an \textit{answer generator}, a \textit{verifier} and a \textit{justification generator}.
The system leverages both an LLM and an MLLM, both based on Gemini-2.0-Flash,
with each model assigned distinct roles at different stages of the pipeline.

Given an image-text claim, the question generator, implemented with Gemini-2.0-Flash, first produces five evidence-seeking questions for claim verification. To enhance question quality, we apply few-shot prompting using annotated questions from the top-3 most similar claims in the training split, where similarity is computed with BM25~\cite{DBLP:journals/ftir/RobertsonZ09} over the textual content of claims.

The answer generator then addresses each generated question by automatically selecting an appropriate answering strategy. For image-related questions that focus on visual cues, Gemini-2.0-Flash is used as a visual question answering (VQA) model. For image-related questions requiring external knowledge, the top-30 pieces of relevant image-related textual evidence are retrieved from the knowledge store using BM25, and Gemini-2.0-Flash generates answers conditioned on the retrieved evidence. For purely textual questions, the system similarly retrieves the top-30 pieces of claim-related textual evidence and produces text-based answers with Gemini-2.0-Flash augmented with the retrieved context.
When an answer is determined to be image-based, the system retrieves the top-2 candidate images from the image evidence set using CLIP-based similarity~\cite{DBLP:conf/icml/RadfordKHRGASAM21} between the textual query and images. Gemini-2.0-Flash then selects the most appropriate image as the final answer via VQA.

After all questions are answered, the verifier, Gemini-2.0-Flash, aggregates the collected evidence and predicts a verdict for the claim. Finally, Gemini-2.0-Flash will serve as the justification generator that produces a natural-language explanation of the predicted verdict based on the pool of evidence (i.e., question-answer pairs).
Additional implementation details and alternative baseline variants are described in the original~\cite{DBLP:journals/corr/abs-2505-17978}.

\subsection{Evaluation}
\label{sec:task-eval}
The evaluation primarily focuses on verdict accuracy conditioned on the quality of the retrieved evidence, termed as the \our score. Specifically, a verdict is considered valid only when its associated evidence score exceeds a predefined threshold.
This design encourages systems not only to produce correct verdicts, but also to retrieve appropriate evidence.
In addition to the \our score, we report auxiliary metrics including the evidence score, question score and the justification score for reference and further analysis.

The evaluation of evidence largely follows the methodology introduced in~\cite{DBLP:journals/corr/abs-2505-17978}. Specifically, it extended Ev2R~\cite{DBLP:journals/corr/abs-2411-05375}, an LLM-as-a-Judge framework for evidence evaluation, to a multimodal setting where evidence may consist of both textual and visual components.
Given a piece of predicted evidence, we conduct reference-based evaluations of its textual and visual parts separately.
For the textual component, we adopt the same evaluation protocol used in the previous shared task~\cite{akhtar-etal-2025-2nd}.
If the textual part of the predicted evidence matches the textual component of the corresponding GT evidence, we further evaluate the visual component by comparing it against the associated images in the GT annotations to assess visual similarity. The visual similarity assessment is formulated as a VQA task, in which the evaluation model assigns a similarity score on a 0-10 scale to a pair of images, with higher scores indicating greater visual similarity. Image pairs receiving a score below 8 are considered insufficiently similar. In such cases, the corresponding evidence match is deemed invalid due to a visual mismatch.

We report evidence \textit{recall}, defined as the percentage of GT evidence instances that are successfully retrieved by the system.
Unlike the original \our paper~\cite{DBLP:journals/corr/abs-2505-17978}which relied on a closed-source model, Gemini, for evaluation, we instead adopt the open-source Gemma-3-27B model~\cite{DBLP:journals/corr/abs-2503-19786} to improve transparency and reproducibility.

For the evaluation of generated questions, we apply Ev2R by directly comparing predicted questions against GT questions.
For justification generation, prior work~\cite{DBLP:journals/corr/abs-2505-17978} showed that ROUGE-1~\cite{lin-2004-rouge} provides a coarse baseline but lacks the flexibility required to assess open-ended generation. Accordingly, in this shared task we adopt a reference-based evaluation using Ev2R, which has demonstrated strong alignment with human judgments for open-ended text generation through comparison with human-annotated references.

Following~\citet{DBLP:journals/corr/abs-2505-17978}, we empirically set the evidence score threshold to 0.3. Claims with evidence scores below this threshold receive an \our score of 0.
Consistent with previous shared tasks~\cite{DBLP:journals/corr/abs-2410-23850,akhtar-etal-2025-2nd}, we limit the maximum number of evidence pieces returned per claim to 10. Additionally, we cap the length of each evidence item at 1,500 tokens, based on empirical estimates of evidence lengths observed in submitted systems.

\begin{table*}[ht]
    \centering
\begin{tabular}{ll|ccc|r}
\toprule
\textbf{Rank} & \textbf{Team Name}                     & \textbf{Ques.} &\textbf{Evid.}  & \textbf{Just.}& \our\\ \midrule
1&HUMANE & 0.8897&	0.5358&	0.5557&	0.5455\\
2&ADA-AGGR&	0.3701&	0.4629	&0.4331&0.5369	 \\
3&AIC CTU&	0.8065&	0.3251		&0.3043&0.3466\\
4&XxP&	0.3902&	0.2703	&	0.1980&0.2557\\
5& REVEAL	&0.6317&	0.2771&0.1348&	0.2358	\\
6&fv	&0.2885	&0.1626	&	0.1306&0.1591\\
\midrule
7&Baseline &0.5545&	0.1707&	0.1322&	0.1136 \\
\bottomrule 
\end{tabular}
    \caption{\textbf{Overall testing results for the shared task.}}
    \label{tab:overall_results}
\end{table*}
\section{Results}
\label{sec:results}
During the development phase, we received submissions from 14 teams. Among them, six teams participated in the testing phase, and five teams submitted system description papers to the workshop. The testing-phase results are reported in Table~\ref{tab:overall_results}, while the core methodological components of the submitted systems are summarized in Table~\ref{tab:component}. We next provide an overview of the main techniques adopted by participating teams.

\begin{table*}[ht]
    \centering
    \rowcolors{2}{white}{gray!15}
\begin{tabular}{l|lp{2.0cm}p{2.8cm}p{2.0cm}lp{2.0cm}}
\toprule
\textbf{Team Name} & \textbf{Evid.} & \textbf{QG} & \textbf{Retrieval} & \textbf{QA} & \textbf{Iter.}& \textbf{Veracity} \\ \midrule
HUMANE& KS$^{\uparrow}$ &Gemini-2.5-Pro&mxbai-embed-largev1, mxbai-rerank-large-v1 &Gemini-2.5-Pro$^{\clubsuit}$&$\checkmark$&Gemini-2.5-Pro\\
ADA-AGGR& KS&Gemini-3-Pro&BM25, SFR-embedding-2, Llama-3.1-70b, ColPali$^\spadesuit$ &Gemini-3-Pro&\ding{56}&Gemini-3-Pro\\
AIC CTU&KS$^{\uparrow}$&GPT-5.1&mxbai-embed-large-v1&GPT-5.1$^{\clubsuit}$&\ding{56}&GPT-5.1\\
Xxp&KS& Qwen3-VL-8B-Instruct &BM25, gte-base, SigLIP$^\spadesuit$&-&\ding{56}&Qwen3-VL-8BInstruct\\
REVEAL&KS&Qwen2.5-VL-7B-Instruct&BM25, SFR-Embedding-2\_R, SigLIP2-Large&Qwen2.5-VL-7B-Instruct&\checkmark&Qwen2.5-VL-7B-Instruct\\
\midrule
Baseline&KS&Gemini-2.0-Flash&BM25&Gemini-2.0-Flash&\ding{56}&Gemini-2.0-Flash\\
\bottomrule 
\end{tabular}
    \caption{\textbf{Summary of essential components of the submitted systems.} KS$^\uparrow$ indicates the use of additional or updated knowledge sources beyond the provided knowledge store. In the QA column, $\clubsuit$ denotes systems that jointly generate questions and answers, while $-$ indicates that no explicit answering stage is involved. In the retrieval column, $\spadesuit$ denotes systems that use the generated questions during the retrieval stage.}
    \label{tab:component}
\end{table*}

\noindent\textbf{Knowledge Source: }All teams built their systems on top of the provided knowledge store, while two teams further updated it. The HUMANE team identified empty entries in the original knowledge store, primarily caused by access-restricted websites (e.g., login walls).
In addition, the basic scraping method employed during knowledge store curation with the trafilatura package occasionally extracted non-informative content that contained only generic website components (e.g., navigation bars, footers, or cookie notices). Such limitations in scraping substantially hinder the acquisition of meaningful information from online resources.
To address this issue, the HUMANE team leveraged Playwright\footnote{https://github.com/microsoft/playwright}, a browser automation framework, to scrape textual content from URLs in the provided knowledge store. By adopting this more advanced scraping strategy, they increased the amount of textual evidence by $24.4\%$ and $15.7\%$ on the test split for claim-text-related and claim-image-related evidence, respectively.

On the other hand, the AIC CTU team focused on improving the collection of claim-image-related evidence. They employed Google Lens\footnote{https://lens.google.com/} as a complementary RIS engine to retrieve web pages associated with the claim image, retaining only those pages that contained images visually similar to the input claim image.
Temporal constraints were also applied during the RIS stage. Subsequently, the Firecrawl API\footnote{https://firecrawl.dev/} was used to scrape textual content from the filtered web pages. By employing multiple RIS engines, their approach mitigated the issue of empty returned web pages related to claim images, a limitation previously identified by work~\cite{DBLP:conf/emnlp/TongletMG24}.

\noindent\textbf{Question Generation: }
Given the multimodal nature of image-text claim verification, all teams employed an MLLM to generate questions relevant to the verification process.
An interesting observation is that some teams (i.e., those \textit{without} $\spadesuit$ in the \textit{Retrieval} column of Table~\ref{tab:component})
did not use the generated questions during evidence retrieval. However, there is a lack of systematic analysis on how different types of textual queries affect retrieval performance. As a result, it remains unclear whether incorporating generated questions into retrieval queries can improve evidence collection for image-text claim verification.

Notably, two teams, HUMANE and AIC CTU, which achieved the highest question generation scores, leveraged training data for few-shot learning. This indicates that few-shot demonstrations can effectively guide models to generate more essential and verification-oriented questions.

\noindent\textbf{Evidence Retrieval: }
The original baseline employed relatively simple retrieval methods, relying on a vanilla coarse-ranking approach using BM25~\cite{DBLP:journals/ftir/RobertsonZ09} for textual evidence and CLIP similarity~\cite{DBLP:conf/icml/RadfordKHRGASAM21} for image evidence. In contrast, all participating teams substantially strengthened the retrieval stage through more carefully designed pipelines.

For textual evidence, inspired by prior findings in text-based claim verification, 
all teams implemented two-stage retrieval frameworks that combine sparse and dense retrieval. 
Specifically, sparse retrievers (e.g., BM25~\cite{DBLP:journals/ftir/RobertsonZ09}) were first employed to efficiently narrow down candidate evidence through lexical matching and precise keyword overlap, ensuring high recall of potentially relevant documents. Dense retrievers were then applied to re-rank or further retrieve semantically relevant evidence using neural embeddings, enabling the systems to capture paraphrased or contextually similar information beyond exact token matches.
Teams consistently adopted more sophisticated embedding models (e.g., mxbai-embed-largev1~\cite{mxbai-embedding}, Llama-3.1-70b~\cite{llama-meta}) to improve evidence dense representations.
The resulting gains in evidence scores highlight the critical role of high-quality evidence and dense retrieval in effective evidence selection.
Furthermore, the REVEAL team demonstrated that generating hypothetical evidence snippets with LLMs can facilitate downstream evidence retrieval, echoing observations from earlier text-only claim verification systems~\cite{yoon-etal-2024-hero}.

Meanwhile, several teams incorporated multimodal retrievers to explicitly account for the multimodal nature of the task.
ADA-AGGR employed the multimodal retriever ColPali~\cite{DBLP:conf/iclr/FaysseSWOVHC25} to both refine textual evidence related to claim images and retrieve image evidence.
Xxp adopted SigLIP~\cite{DBLP:conf/iccv/ZhaiM0B23}, an improved variant of CLIP~\cite{DBLP:conf/icml/RadfordKHRGASAM21}, to enhance the retrieval of textual evidence associated with claim images.
In contrast, REVEAL utilized SigLIP2~\cite{DBLP:journals/corr/abs-2502-14786} select the most relevant image evidence given a textual query.
Both ADA-AGGR and REVEAL conducted ablation studies, demonstrating that incorporating multimodal retrievers leads to measurable performance gains. Their results further indicate that the choice and quality of the multimodal retriever play a crucial role in fact-checking performance on image-text claims.

\begin{table}[ht]
    \centering
\begin{tabular}{l|ccc}
\toprule
 \textbf{Team Name} & \textbf{Before} & \textbf{After} & \textbf{Overall}\\
 \midrule
HUMANE &0.5333 &0.7273 &0.5455 \\
ADA-AGGR &0.5303 &0.6364 & 0.5369\\
AIC CTU &0.3333& 0.5455&0.3466 \\
XxP &0.2545 &0.2727 &0.2557 \\
REVEAL	& 0.2424& 0.1364& 0.2358\\
fv	&0.1485 &0.2727 & 0.1591\\
\midrule
Baseline &0.1091 &0.1818&0.1136 \\
\bottomrule 
\end{tabular}
    \caption{\textbf{\our scores on 330 test claims published before and 22  published after January 2025.}}
    \label{tab:cutoff_results}
\end{table}

\begin{table*}[ht]
    \centering
\begin{tabular}{l|cccc|cccc|r}
\toprule
 \textbf{Team Name} 
 & \textbf{EP} & \textbf{MA} & \textbf{Cs} & \textbf{Nm}
 & \textbf{S} & \textbf{R} & \textbf{NEE} & \textbf{CE/C}
 & \textbf{Avg. \# Docs}\\ \midrule
HUMANE & 0.54&0.57 &0.83 &0.80 & 0.65&0.57 & 0.00&0.29 &  10.0\\
ADA-AGGR &0.54 & 0.52& 0.67& 0.20& 0.63& 0.57&0.05 &0.00 &11.2  \\
AIC CTU &0.34 &0.37 & 0.42& 0.40&0.51 & 0.34&0.14 &0.00 &  9.78\\
XxP &0.26 & 0.25& 0.17& 0.60& 0.24& 0.28& 0.00&0.00 & 7.50 \\
REVEAL	&0.24 & 0.14&0.08 &0.00 &0.00 &0.30 &0.05 & 0.00& 7.63 \\
fv	&0.16 & 0.18&0.17 &0.20 & 0.18& 0.17& 0.00&0.00 & 3.00 \\
\midrule
Baseline &0.11 &0.14 &0.08 &0.00 &0.29& 0.08& 0.19& 0.00& 5.00 \\
\bottomrule 
\end{tabular}
    \caption{\textbf{\our scores of different claim types and verdict types.} We provide results over four most frequent claim types (EP = Event/Property, MA = Media Analysis, Cs = Causal, Nm = Numerical).
    Results over different verdict types (S =Supported, R =Refuted, NEE = Not Enough Evidence, CE/C = Conflicting Evidence / Cherrypicking) are reported.
    We also report the average number of pieces of evidence per team.
    We note that if a team submitted more than 10 pieces of evidence for a claim, only the first 10 were considered for evaluation.}
    \label{tab:indv_class_results}
\end{table*}

\noindent\textbf{Question Answering: }
As identified in the \our paper, fact-checking often involves dependencies across reasoning steps, where the generation of subsequent questions may rely on evidence retrieved from earlier QA pairs.
Two teams, HUMANE and REVEAL, explicitly modeled this dependency by iteratively generating QA pairs and questions conditioned on previous QA history.
However, none of the submitted systems conducted controlled experiments comparing iterative QA generation with one-time question generation, leaving the impact of this design choice underexplored.

Interestingly, the team Xxp generated questions but did not answer them. Instead, they treated the evidence retrieved using the claim
as the final predicted evidence.
ADA-AGGR employed question answering only as a mechanism for refining image evidence, specifically for selecting the most relevant images.
Their ablation study showed a significant reduction in computation time when the QA module was removed, with little performance degradation.
These observations raise an open question: should retrieved evidence primarily be used to answer information-seeking questions as an additional refinement step, or is retrieval alone sufficient and precise enough for effective claim verification?

HUMANE and AIC CTU further combined question generation and question answering into a single step. In their question-answer generation step, the claims were fed into MLLMs. 
Although these systems incorporated retrieved evidence, it remains unclear whether the generated QA pairs relied on the models’ parametric knowledge or on external evidence.
To investigate this issue, we split the test set into claims published before and after January 2025, corresponding to the knowledge cutoff of Gemini-2.5-Pro. 
As shown in Table~\ref{tab:cutoff_results}, most teams achieved higher performance on more recent claims, consistent with observations from previous shared tasks~\cite{akhtar-etal-2025-2nd,DBLP:journals/corr/abs-2410-23850}. This trend likely reflects the greater availability and retrievability of evidence for recent events.
These results highlight the importance of more rigorously isolating parametric knowledge from retrieved evidence in order to assess models’ true fact-checking capabilities.


\noindent\textbf{Verdict Prediction: }
All participating teams employed MLLMs for verdict prediction. Notably, instead of assessing the stance of each retrieved evidence item individually~\cite{DBLP:conf/nips/SchlichtkrullG023}, all systems fed the complete set of retrieved evidence jointly into the verdict prediction model. This design choice is consistent with findings from the \our paper, which emphasize the necessity of joint reasoning over multiple evidence pieces for image-text claim verification.
Furthermore, all teams explicitly modeled all four verdict classes, including infrequent types.
Among them,
ADA-AGGR and Xxp additionally incorporated explanations of each verdict category into their prompts, providing clearer guidance to the model during prediction.

\noindent\textbf{Performance across Claim Types and Verdict Types: }
Table~\ref{tab:indv_class_results} reports results across different claim types (event/property, media analysis, causal and numerical) as well as verdict types (supported, refuted, not enough evidence and conflicting evidence / cherrypicking). 
We observe that the two top-performing teams substantially outperform the remaining systems on event/property and media analysis claims, which constitute the majority of the dataset.
Lower-performing systems struggle with causal claims, a trend that is consistent with findings from previous shared tasks~\cite{akhtar-etal-2025-2nd,DBLP:journals/corr/abs-2410-23850}.
However, the results may not fully reflect model performance on causal and numerical claims due to the limited number of such claims in the testing split.

Across verdict types, all systems exhibit notably poor performance on NEE and CE/C claims, with \our scores close to zero.
This result can be attributed, in part, to the limited number of instances in these categories, which restricts reliable evaluation. More fundamentally, modeling conflicting or insufficient evidence poses intrinsic challenges.
Specifically, recent studies have shown that existing foundation models are considerably less capable of reasoning under conflicting contexts~\cite{DBLP:conf/naacl/LiuNHQXHZJMBR25,DBLP:conf/ijcai/GeWCL025}.
These findings highlight the need for more sophisticated system designs that explicitly address evidence conflict and uncertainty.

\begin{table}[t]
    \centering
    \begin{tabular}{l|cc|cc}
    \toprule
    & \multicolumn{2}{c|}{Human-Human} & \multicolumn{2}{|c}{Human-Model}\\
    \midrule
     \textbf{Dimension} & \textbf{Cov.} & \textbf{Rele.} & \textbf{Cov.} & \textbf{Rele.} \\
    \midrule
     $\rho$ & 0.499& 0.529  & 0.310 &0.319\\
     $r$ & 0.516 & 0.578   & 0.325 &0.355\\
    \bottomrule
    \end{tabular}
    \caption{\textbf{Correlation between human annotators and between our evidence evaluation scores and human-rated scores.} 
    We calculate correlation using the Spearman ($\rho$) and Pearson ($r$) correlation coefficients.}
    \label{tab:correlation_metrics}
\end{table}

\section{Human Evaluation of Evidence}
\label{sec:human-eval}
Following previous shared tasks~\cite{DBLP:journals/corr/abs-2410-23850,akhtar-etal-2025-2nd}, we conducted a human evaluation of the evidence retrieved by participating systems. In the original \our paper, a human alignment check was performed to validate the reliability of the automatic evaluation method by comparing two sets of human-annotated evidence, treating one as the reference. 
We acknowledge that discrepancies may exist between human-annotated evidence and model-predicted evidence. Leveraging the availability of diverse predicted evidence from multiple systems in this shared task, we therefore extend this analysis by assessing the alignment between the automatic evaluation method and human judgments, through direct comparison of models’ predicted evidence with reference (human-annotated) evidence.

\noindent\textbf{Evaluation Process: }
We conducted a human evaluation of predicted evidence in collaboration with participants of the shared task. All five teams that submitted shared task papers participated in the evaluation. 
We collected predicted evidence from different systems. 
Specifically, predicted evidence for 25 claims was collected from each team.
To ensure the robustness of the evaluation, evidence samples were randomly selected across systems, with their automatic evaluation scores uniformly distributed between 0 and 1. Each predicted evidence sample was independently evaluated by two teams.

For each instance, we provided annotators with the original claim (including claim text and associated image(s)), relevant metadata, the predicted evidence, and the reference (human-annotated) evidence. Following the human evaluation protocols adopted in previous shared tasks, annotators were asked to rate the predicted evidence along two dimensions, \textit{coverage} and \textit{relevance}, by comparison with the reference evidence.
Coverage measures the extent to which the predicted evidence fully captures the content of the reference evidence, including its meaning, entities, and other key informational elements. Given the multimodal nature of the task, annotators were explicitly instructed that if the image component of a reference evidence item is not reflected in the corresponding predicted evidence, the coverage score should be reduced.
Relevance, assesses how useful the predicted evidence is for verifying the claim.

Annotators rated each predicted evidence sample on a scale from 0 to 5 for both dimensions. Detailed human evaluation guidelines are provided in Appendix~\ref{sec:app-human}.

\noindent\textbf{Evaluation Results: }
The annotation process resulted in 250 human annotations on evidence predictions for 100 claims that each was annotated by two different participants. We report the correlation between human ratings and automatic evidence evaluation using Spearman ($\rho$)~\citep{spearman04} and Pearson ($r$)~\citep{pearson1896mathematical} correlation coefficient. 
The results, shown in Table~\ref{tab:correlation_metrics}, suggests that while human annotators exhibit relatively high agreement with each other in evidence evaluation, the automatic evaluation method shows limited alignment with human judgments.

To better understand the sources of this misalignment, we conducted a fine-grained analysis across predictions from different teams (full results are reported in Appendix~\ref{sec:app-human}).
We observe that the automatic evaluation model performs particularly poorly on predictions from the HUMANE team, yielding a Spearman correlation coefficient of $0.06$ and a Pearson correlation coefficient of $0.04$ regarding coverage.
The HUMANE submission consists exclusively of text-only evidence, whereas the reference evidence is multimodal in some cases, incorporating both the claim image and associated external information. 
Although annotators were instructed to consider image information during evidence evaluation, participants often judged the inclusion of claim images as unnecessary and the evidence text alone was deemed sufficient, as no additional external visual evidence was required.
This discrepancy appears to negatively affect automatic evaluation, which relies on reference evidence that may explicitly encode multimodal information.
In contrast, the automatic evaluation of the REVEAL team’s predicted evidence exhibits substantially stronger alignment with human assessments on coverage, with $\rho = 0.575$ and $r = 0.520$. 

We further observe lower agreement among human annotators when evaluating the coverage of HUMANE’s predicted evidence ($\rho = 0.483$ and $r = 0.655$), suggesting intrinsic ambiguity in judging text-only evidence against multimodal references.
These findings highlight a key challenge for automatic evidence evaluation in image-text claim verification: determining whether and how claim images should be explicitly represented and accounted for in the evidence set.

At the same time, we conduct an alignment analysis on predicted evidence with the lowest and highest automatic evaluation scores.
We observe that the automatic evaluation method 
yields slightly more reliable 
when assigning higher scores,
achieving higher correlation with human ratings ($\rho = 0.347$, $r = 0.370$).
This observation also helps explain why higher agreement was reported in the original \our paper.
In that setting, the evaluation method was applied to comparatively high-quality evidence annotated by humans and evaluated against annotations provided by another annotator.
Overall, human assessments of predicted evidence largely align with the ranking of participating systems (details in Appendix~\ref{sec:app-human}).

\section{Lessons Learned}
\label{sec:lessons}
From this shared task, we derive three key insights.
First, more robust scraping methods are required. In the originally provided knowledge store, some entries were empty due to website inaccessibility, while others contained limited or low-quality information. As demonstrated by the HUMANE team, employing more sophisticated scraping tools can substantially improve the informativeness of the collected content.

Second, the use of multiple RIS engines is crucial, as it increases the coverage of web pages associated with a given claim image. In practice, a single RIS engine may fail to retrieve any relevant pages, whereas combining multiple engines can effectively mitigate this limitation.

Third, we observe that top-performing teams rely heavily on closed-source models. While the ADA-AGGR team explored fine-tuning open-weight models and demonstrated that performance gains are possible, closed-source models still exhibit a clear advantage. Nevertheless, open-weight models offer greater transparency and can significantly reduce the cost associated with API-based closed-source systems. How to more effectively leverage training data and fine-tune open-weights models to narrow this performance gap remains an underexplored and important research direction.

\section{Conclusions \& Future Work}
In the shared task, all teams have outperformed our baseline. We analyzed key components of different systems and their association with the end performance, highlighting key takeaways from the shared task.
We observed that the top performing teams are heavily relying on closed-source models. In the future, it is worth exploring whether using open-source models and training data properly could bridge the performance gap between the performance with closed-source ones. 
Meanwhile, there are three open questions: 1) whether generating information-seeking queries are needed for claim verification, 2) whether we should further refine retrieved evidence with the QA step and 3) whether iterative QA is helpful. By answering these questions, we can remove unnecessary modules of systems and seek a balance between effectiveness and efficiency.
Further improvement over the automatic evidence evaluation method is also needed to better align with human judgment. 

\section{Limitations \& Ethics}
The claims in \our are derived from fact-checking articles. As a result, the dataset may inherit biases inherent in these sources, including selection bias~\cite{10.1111/jcom.12284,Barnoy10122019}. Moreover, the dataset and associated models are not designed for absolute truth discovery. Instead, the veracity labels in \our are conditioned on the evidence retrieved by annotators and therefore reflect the perspectives and biases of both annotators and journalists. Consequently, participating systems optimized for performance on \our, may replicate these biases.

Considerable efforts were made to mitigate temporal leakage by enforcing that all evidence must be published prior to the date of the associated claim. However, this constraint does not fully eliminate leakage at the model level, as foundation models may have already encountered these claims during pre-training. For example, the cutoff date of Gemini-2.5-Pro is January 2025, whereas only 22 test claims were published after this date.

Finally, while reference-based evaluation is effective for assessing textual evidence, we observe that it is substantially less reliable for evaluating evidence in image-text claims. The current evaluation method exhibits limited alignment with human judgments, highlighting the need for more robust evaluation strategies tailored to multimodal evidence.

\section*{Acknowledgments}
Rui and Andreas were funded by a grant from the Alan Turing Institute and DSO National Laboratories (Singapore). 
Zhenyun, Yulong, Michael, and Andreas received funding from the European Research Council (ERC) under the European Union’s Horizon 2020 Research and Innovation programme grant AVeriTeC (Grant agreement No. 865958).
Rui and Andreas Vlachos receive further support from the DARPA program SciFy.
We would like to thank participants from the shared task (Max Upravitelev, Herbert Ullrich, Yoana Tsoneva, and Amina Tariq) for contributing to the human evaluation of evidence.
\bibliography{anthology,custom}

@inproceedings{DBLP:conf/nips/SchlichtkrullG023,
  author       = {Michael Sejr Schlichtkrull and
                  Zhijiang Guo and
                  Andreas Vlachos},
  title        = {AVeriTeC: {A} Dataset for Real-world Claim Verification with Evidence
                  from the Web},
  booktitle    = {Advances in Neural Information Processing Systems 36: Annual Conference
                  on Neural Information Processing Systems 2023, NeurIPS} ,
  year         = {2023},
  volume={36},
  pages={65128--65167},
}

@inproceedings{DBLP:conf/naacl/LiuNHQXHZJMBR25,
  author       = {Siyi Liu and
                  Qiang Ning and
                  Kishaloy Halder and
                  Zheng Qi and
                  Wei Xiao and
                  Phu Mon Htut and
                  Yi Zhang and
                  Neha Anna John and
                  Bonan Min and
                  Yassine Benajiba and
                  Dan Roth},
  title        = {Open Domain Question Answering with Conflicting Contexts},
  booktitle    = {Findings of the Association for Computational Linguistics: {NAACL}},
  series       = {Findings of {ACL}},
  pages        = {1838--1854},
  year         = {2025}
}

@inproceedings{DBLP:conf/ijcai/GeWCL025,
  author       = {Ziyu Ge and
                  Yuhao Wu and
                  Daniel Wai Kit Chin and
                  Roy Ka{-}Wei Lee and
                  Rui Cao},
  title        = {Resolving Conflicting Evidence in Automated Fact-Checking: {A} Study
                  on Retrieval-Augmented LLMs},
  booktitle    = {Proceedings of the Thirty-Fourth International Joint Conference on
                  Artificial Intelligence, {IJCAI}},
  pages        = {9656--9664},
  year         = {2025}
}

@inproceedings{DBLP:conf/naacl/ThorneVCM18,
  author       = {James Thorne and
                  Andreas Vlachos and
                  Christos Christodoulopoulos and
                  Arpit Mittal},
  title        = {{FEVER:} a Large-scale Dataset for Fact Extraction and VERification},
  booktitle    = {Proceedings of the 2018 Conference of the North American Chapter of
                  the Association for Computational Linguistics: Human Language Technologies,
                  {NAACL-HLT}},
  pages        = {809--819},
  year         = {2018}
}

@inproceedings{DBLP:conf/naacl/ChenKSDC24,
  author       = {Jifan Chen and
                  Grace Kim and
                  Aniruddh Sriram and
                  Greg Durrett and
                  Eunsol Choi},
  title        = {Complex Claim Verification with Evidence Retrieved in the Wild},
  booktitle    = {Proceedings of the 2024 Conference of the North American Chapter of
                  the Association for Computational Linguistics: Human Language Technologies
                  (Volume 1: Long Papers), {NAACL}},
  pages        = {3569--3587},
  year         = {2024}
}

@inproceedings{DBLP:conf/nips/AlyGST00CM21,
  author       = {Rami Aly and
                  Zhijiang Guo and
                  Michael Sejr Schlichtkrull and
                  James Thorne and
                  Andreas Vlachos and
                  Christos Christodoulopoulos and
                  Oana Cocarascu and
                  Arpit Mittal},
  title        = {{FEVEROUS:} Fact Extraction and VERification Over Unstructured and
                  Structured information},
  booktitle    = {Proceedings of the Neural Information Processing Systems Track on
                  Datasets and Benchmarks 1, NeurIPS Datasets and Benchmarks},
  year         = {2021}
}

@inproceedings{DBLP:conf/emnlp/AlhindiPM18,
  author       = {Tariq Alhindi and
                  Savvas Petridis and
                  Smaranda Muresan},
  title        = {Where is Your Evidence: Improving Fact-checking by Justification Modeling},
  booktitle    = {Proceedings of the First Workshop on Fact Extraction and VERification,
                  FEVER@EMNLP},
  pages        = {85--90},
  year         = {2018}
}

@inproceedings{DBLP:conf/sigir/YaoS0CH23,
  author       = {Barry Menglong Yao and
                  Aditya Shah and
                  Lichao Sun and
                  Jin{-}Hee Cho and
                  Lifu Huang},
  title        = {End-to-End Multimodal Fact-Checking and Explanation Generation: {A}
                  Challenging Dataset and Models},
  booktitle    = {Proceedings of the 46th International {ACM} {SIGIR} Conference on
                  Research and Development in Information Retrieval, {SIGIR}},
  pages        = {2733--2743},
  year         = {2023}
}

@article{DBLP:journals/ftir/RobertsonZ09,
  author       = {Stephen E. Robertson and
                  Hugo Zaragoza},
  title        = {The Probabilistic Relevance Framework: {BM25} and Beyond},
  journal      = {Found. Trends Inf. Retr.},
  volume       = {3},
  number       = {4},
  pages        = {333--389},
  year         = {2009}
}

@inproceedings{DBLP:conf/emnlp/LuoDR21,
  author       = {Grace Luo and
                  Trevor Darrell and
                  Anna Rohrbach},
  title        = {NewsCLIPpings: Automatic Generation of Out-of-Context Multimodal Media},
  booktitle    = {Proceedings of the 2021 Conference on Empirical Methods in Natural
                  Language Processing, {EMNLP}},
  pages        = {6801--6817},
  year         = {2021}
}

@inproceedings{DBLP:conf/cvpr/JiaHZJCL23,
  author       = {Shan Jia and
                  Mingzhen Huang and
                  Zhou Zhou and
                  Yan Ju and
                  Jialing Cai and
                  Siwei Lyu},
  title        = {AutoSplice: {A} Text-prompt Manipulated Image Dataset for Media Forensics},
  booktitle    = {{IEEE/CVF} Conference on Computer Vision and Pattern Recognition,
                  {CVPR}},
  pages        = {893--903},
  year         = {2023}
}

@article{DBLP:journals/corr/abs-2405-11697,
  author       = {Nicholas Dufour and
                  Arkanath Pathak and
                  Pouya Samangouei and
                  Nikki Hariri and
                  Shashi Deshetti and
                  Andrew Dudfield and
                  Christopher Guess and
                  Pablo Hern{\'{a}}ndez Escayola and
                  Bobby Tran and
                  Mevan Babakar and
                  Christoph Bregler},
  title        = {AMMeBa: {A} Large-Scale Survey and Dataset of Media-Based Misinformation
                  In-The-Wild},
  journal      = {CoRR},
  volume       = {abs/2405.11697},
  year         = {2024}
}

@article{Newman2012NonprobativeP,
  title={Nonprobative photographs (or words) inflate truthiness},
  author={Eryn J. Newman and Maryanne Garry and Daniel M. Bernstein and Justin Kantner and Stephen Lindsay},
  journal={Psychonomic Bulletin \& Review},
  year={2012},
  volume={19},
  pages={969-974}
}

@inproceedings{DBLP:journals/corr/abs-2407-13488,
    author    = {Papadopoulos, Stefanos-Iordanis and Koutlis, Christos and Papadopoulos, Symeon and Petrantonakis, Panagiotis C.},
    title     = {Similarity over Factuality: Are we Making Progress on Multimodal Out-of-Context Misinformation Detection?},
    booktitle = {Proceedings of the Winter Conference on Applications of Computer Vision (WACV)},
    year      = {2025},
    pages     = {5570-5579}
}

@article{DBLP:journals/ijmir/PapadopoulosKPP24,
  author       = {Stefanos{-}Iordanis Papadopoulos and
                  Christos Koutlis and
                  Symeon Papadopoulos and
                  Panagiotis C. Petrantonakis},
  title        = {{VERITE:} a Robust benchmark for multimodal misinformation detection
                  accounting for unimodal bias},
  journal      = {Int. J. Multim. Inf. Retr.},
  volume       = {13},
  number       = {1},
  pages        = {4},
  year         = {2024}
}

@article{DBLP:journals/corr/abs-2411-05375,
  author       = {Mubashara Akhtar and
                  Michael Sejr Schlichtkrull and
                  Andreas Vlachos},
  title        = {Ev2R: Evaluating Evidence Retrieval in Automated Fact-Checking},
  journal      = {CoRR},
  volume       = {abs/2411.05375},
  year         = {2024}
}

@inproceedings{DBLP:conf/emnlp/OusidhoumY022,
  author       = {Nedjma Ousidhoum and
                  Zhangdie Yuan and
                  Andreas Vlachos},
  title        = {Varifocal Question Generation for Fact-checking},
  booktitle    = {Proceedings of the 2022 Conference on Empirical Methods in Natural
                  Language Processing, {EMNLP}},
  pages        = {2532--2544},
  year         = {2022}
}

@article{spearman04,
  author = {Spearman, C.},
  journal = {American Journal of Psychology},
  pages = {88--103},
  title = {The Proof and Measurement of Association Between Two Things},
  volume = 15,
  year = 1904
}

@article{pearson1896mathematical,
  author = {Pearson, Karl and Henrici, Olaus Magnus Friedrich Erdmann},
  journal = {Philosophical Transactions of the Royal Society of London. Series A, Containing Papers of a Mathematical or Physical Character},
  pages = {253-318},
  title = {VII. Mathematical contributions to the theory of evolution.\& III. Regression, heredity, and panmixia},
  volume = 187,
  year = 1896
}

@article{DBLP:journals/corr/abs-2502-14786,
  author       = {Michael Tschannen and
                  Alexey A. Gritsenko and
                  Xiao Wang and
                  Muhammad Ferjad Naeem and
                  Ibrahim Alabdulmohsin and
                  Nikhil Parthasarathy and
                  Talfan Evans and
                  Lucas Beyer and
                  Ye Xia and
                  Basil Mustafa and
                  Olivier J. H{\'{e}}naff and
                  Jeremiah Harmsen and
                  Andreas Steiner and
                  Xiaohua Zhai},
  title        = {SigLIP 2: Multilingual Vision-Language Encoders with Improved Semantic
                  Understanding, Localization, and Dense Features},
  journal      = {CoRR},
  volume       = {abs/2502.14786},
  year         = {2025}
}

@inproceedings{DBLP:conf/iccv/ZhaiM0B23,
  author       = {Xiaohua Zhai and
                  Basil Mustafa and
                  Alexander Kolesnikov and
                  Lucas Beyer},
  title        = {Sigmoid Loss for Language Image Pre-Training},
  booktitle    = {{IEEE/CVF} International Conference on Computer Vision, {ICCV}},
  pages        = {11941--11952},
  year         = {2023}
}

@inproceedings{DBLP:conf/icml/RadfordKHRGASAM21,
  author       = {Alec Radford and
                  Jong Wook Kim and
                  Chris Hallacy and
                  Aditya Ramesh and
                  Gabriel Goh and
                  Sandhini Agarwal and
                  Girish Sastry and
                  Amanda Askell and
                  Pamela Mishkin and
                  Jack Clark and
                  Gretchen Krueger and
                  Ilya Sutskever},
  title        = {Learning Transferable Visual Models From Natural Language Supervision},
  booktitle    = {Proceedings of the 38th International Conference on Machine Learning,
                  {ICML}},
  volume       = {139},
  pages        = {8748--8763},
  year         = {2021}
}

@misc{llama-meta,
title={Introducing Llama 3.1: Our most capable models to date},
url={https://ai.meta.com/blog/meta-llama-3-1/}, 
author={Meta},
year         = {2024}
}

@misc{mxbai-embedding,
title={Open Source Strikes Bread - New Fluffy Embedding Model},
url={https://www.mixedbread.com/blog/mxbai-embed-large-v1}, 
author={Sean Lee and Aamir Shakir and Darius Koenig and Julius Lipp},
year         = {2024}
}

@inproceedings{yoon-etal-2024-hero,
    title = "{H}er{O} at {AV}eri{T}e{C}: The Herd of Open Large Language Models for Verifying Real-World Claims",
    author = "Yoon, Yejun  and
      Jung, Jaeyoon  and
      Yoon, Seunghyun  and
      Park, Kunwoo",
    booktitle = "Proceedings of the Seventh Fact Extraction and VERification Workshop (FEVER)",
    year = "2024",
    pages = "130--136"
}

@inproceedings{DBLP:conf/iclr/FaysseSWOVHC25,
  author       = {Manuel Faysse and
                  Hugues Sibille and
                  Tony Wu and
                  Bilel Omrani and
                  Gautier Viaud and
                  C{\'{e}}line Hudelot and
                  Pierre Colombo},
  title        = {ColPali: Efficient Document Retrieval with Vision Language Models},
  booktitle    = {The Thirteenth International Conference on Learning Representations,
                  {ICLR}},
  year         = {2025}
}

@article{DBLP:journals/corr/abs-2503-19786,
  author       = {Aishwarya Kamath and
                  Johan Ferret and
                  Shreya Pathak and
                  Nino Vieillard and
                  Ramona Merhej and
                  Sarah Perrin and
                  Tatiana Matejovicova and
                  Alexandre Ram{\'{e}} and
                  Morgane Rivi{\`{e}}re and
                  Louis Rouillard and
                  Thomas Mesnard and
                  Geoffrey Cideron and
                  Jean{-}Bastien Grill and
                  Sabela Ramos and
                  Edouard Yvinec and
                  Michelle Casbon and
                  Etienne Pot and
                  Ivo Penchev and
                  Ga{\"{e}}l Liu and
                  Francesco Visin and
                  Kathleen Kenealy and
                  Lucas Beyer and
                  Xiaohai Zhai and
                  Anton Tsitsulin and
                  R{\'{o}}bert Busa{-}Fekete and
                  Alex Feng and
                  Noveen Sachdeva and
                  Benjamin Coleman and
                  Yi Gao and
                  Basil Mustafa and
                  Iain Barr and
                  Emilio Parisotto and
                  David Tian and
                  Matan Eyal and
                  Colin Cherry and
                  Jan{-}Thorsten Peter and
                  Danila Sinopalnikov and
                  Surya Bhupatiraju and
                  Rishabh Agarwal and
                  Mehran Kazemi and
                  Dan Malkin and
                  Ravin Kumar and
                  David Vilar and
                  Idan Brusilovsky and
                  Jiaming Luo and
                  Andreas Steiner and
                  Abe Friesen and
                  Abhanshu Sharma and
                  Abheesht Sharma and
                  Adi Mayrav Gilady and
                  Adrian Goedeckemeyer and
                  Alaa Saade and
                  Alexander Kolesnikov and
                  Alexei Bendebury and
                  Alvin Abdagic and
                  Amit Vadi and
                  Andr{\'{a}}s Gy{\"{o}}rgy and
                  Andr{\'{e}} Susano Pinto and
                  Anil Das and
                  Ankur Bapna and
                  Antoine Miech and
                  Antoine Yang and
                  Antonia Paterson and
                  Ashish Shenoy and
                  Ayan Chakrabarti and
                  Bilal Piot and
                  Bo Wu and
                  Bobak Shahriari and
                  Bryce Petrini and
                  Charlie Chen and
                  Charline Le Lan and
                  Christopher A. Choquette{-}Choo and
                  CJ Carey and
                  Cormac Brick and
                  Daniel Deutsch and
                  Danielle Eisenbud and
                  Dee Cattle and
                  Derek Cheng and
                  Dimitris Paparas and
                  Divyashree Shivakumar Sreepathihalli and
                  Doug Reid and
                  Dustin Tran and
                  Dustin Zelle and
                  Eric Noland and
                  Erwin Huizenga and
                  Eugene Kharitonov and
                  Frederick Liu and
                  Gagik Amirkhanyan and
                  Glenn Cameron and
                  Hadi Hashemi and
                  Hanna Klimczak{-}Plucinska and
                  Harman Singh and
                  Harsh Mehta and
                  Harshal Tushar Lehri and
                  Hussein Hazimeh and
                  Ian Ballantyne and
                  Idan Szpektor and
                  Ivan Nardini},
  title        = {Gemma 3 Technical Report},
  journal      = {CoRR},
  volume       = {abs/2503.19786},
  year         = {2025}
}

@article{10.1111/jcom.12284,
    author = {Shin, Jieun and Thorson, Kjerstin},
    title = {Partisan Selective Sharing: The Biased Diffusion of Fact-Checking Messages on Social Media},
    journal = {Journal of Communication},
    volume = {67},
    number = {2},
    pages = {233-255},
    year = {2017},
    issn = {0021-9916}
}

@article{Barnoy10122019,
author = {Aviv Barnoy and Zvi Reich},
title = {The When, Why, How and So-What of Verifications},
journal = {Journalism Studies},
volume = {20},
number = {16},
pages = {2312--2330},
year = {2019}
}

@article{doi:10.1177/0022243719881113,
author = {Yiyi Li and Ying Xie},
title ={Is a Picture Worth a Thousand Words? An Empirical Study of Image Content and Social Media Engagement},
journal = {Journal of Marketing Research},
volume = {57},
number = {1},
pages = {1-19},
year = {2020},
doi = {10.1177/0022243719881113},
}

@inproceedings{DBLP:conf/emnlp/TongletMG24,
  author       = {Jonathan Tonglet and
                  Marie{-}Francine Moens and
                  Iryna Gurevych},
  title        = {"Image, Tell me your story!" Predicting the original meta-context
                  of visual misinformation},
  booktitle    = {Proceedings of the 2024 Conference on Empirical Methods in Natural
                  Language Processing, {EMNLP} },
  pages        = {7845--7864},
  year         = {2024}
}

@article{DBLP:journals/corr/abs-2505-17978,
  author       = {Rui Cao and
                  Zifeng Ding and
                  Zhijiang Guo and
                  Michael Sejr Schlichtkrull and
                  Andreas Vlachos},
  title        = {AVerImaTeC: {A} Dataset for Automatic Verification of Image-Text Claims
                  with Evidence from the Web},
  journal      = {CoRR},
  volume       = {abs/2505.17978},
  year         = {2025}
}

@article{DBLP:journals/corr/abs-2410-23850,
  author       = {Michael Sejr Schlichtkrull and
                  Yulong Chen and
                  Chenxi Whitehouse and
                  Zhenyun Deng and
                  Mubashara Akhtar and
                  Rami Aly and
                  Zhijiang Guo and
                  Christos Christodoulopoulos and
                  Oana Cocarascu and
                  Arpit Mittal and
                  James Thorne and
                  Andreas Vlachos},
  title        = {The Automated Verification of Textual Claims (AVeriTeC) Shared Task},
  journal      = {CoRR},
  volume       = {abs/2410.23850},
  year         = {2024}
}

@inproceedings{akhtar-etal-2025-2nd,
    title = "The 2nd Automated Verification of Textual Claims ({AV}eri{T}e{C}) Shared Task: Open-weights, Reproducible and Efficient Systems",
    author = "Akhtar, Mubashara  and
      Aly, Rami  and
      Chen, Yulong  and
      Deng, Zhenyun  and
      Schlichtkrull, Michael  and
      Whitehouse, Chenxi  and
      Vlachos, Andreas",
    booktitle = "Proceedings of the Eighth Fact Extraction and VERification Workshop (FEVER)",
    year = "2025",
    pages = "201--223"
}

@inproceedings{DBLP:conf/acl/Barbaresi21,
  author       = {Adrien Barbaresi},
  title        = {Trafilatura: {A} Web Scraping Library and Command-Line Tool for Text
                  Discovery and Extraction},
  booktitle    = {Proceedings of the Joint Conference of the 59th Annual Meeting of
                  the Association for Computational Linguistics and the 11th International
                  Joint Conference on Natural Language Processing, {ACL}},
  pages        = {122--131},
  year         = {2021}
}

\clearpage
\appendix

\section{Queries for Knowledge Store Construction}
\label{sec:app-ks}
To simulate realistic evidence retrieval conditions and introduce both diversity and noise into the evidence pool, we augment search queries conditioned on the claim as well as on ground-truth annotated textual questions. In addition, we construct adversarial queries by perturbing key entities, dates, and events mentioned in the claim. This query augmentation strategy is designed to better reflect real-world retrieval scenarios and challenge downstream verification models. The full list of textual query types used for knowledge store construction is provided in Table~\ref{table:knowledge_store_prompts}.

\begin{figure}[ht!]
     \centering
     \includegraphics[scale=0.7]{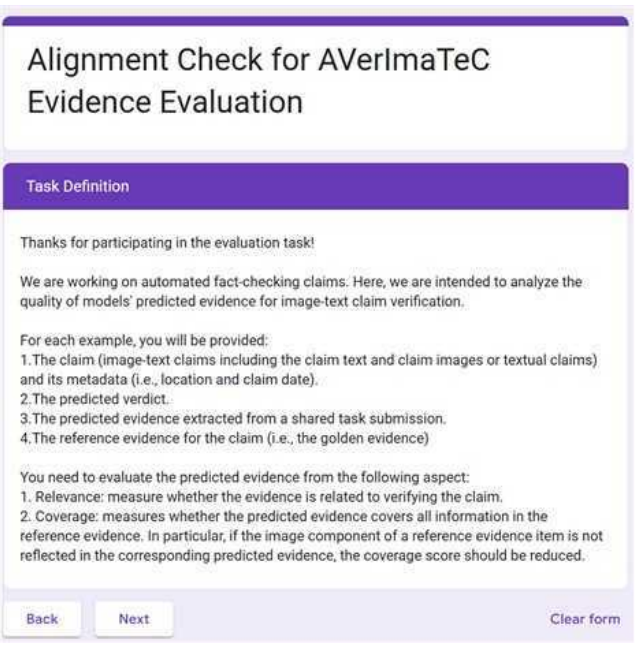}
     \caption{\textbf{Annotation guidelines for human evaluation of predicted evidence.}}
     \label{fig:anno-guideline}
 \end{figure}
 
 \begin{figure}[ht!]
     \centering
     \includegraphics[scale=0.75]{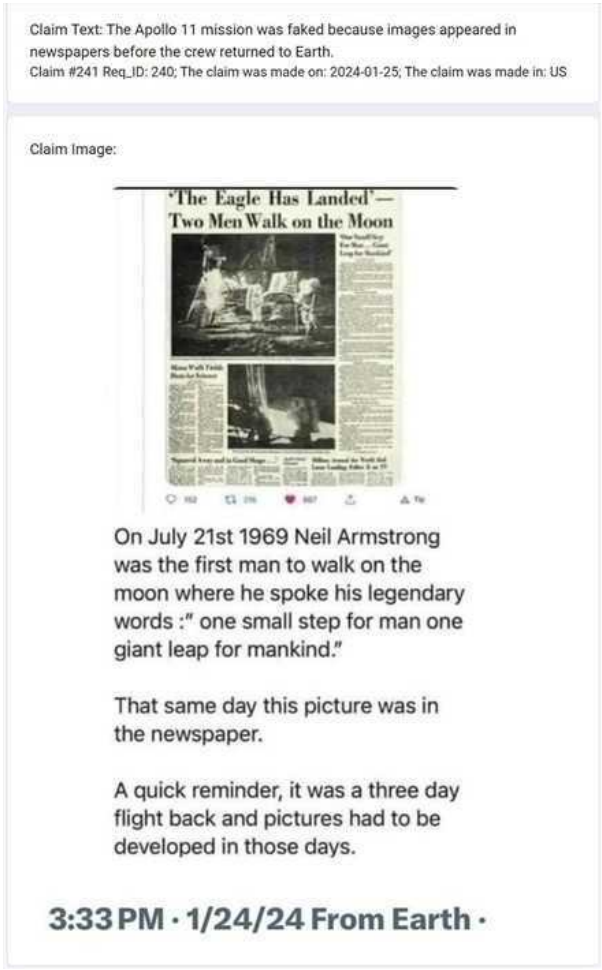}
     \caption{\textbf{An example claim presented to annotators during evaluation.}}
     \label{fig:anno-claim}
 \end{figure}

 \begin{figure}[ht!]
     \centering
     \includegraphics[scale=0.58]{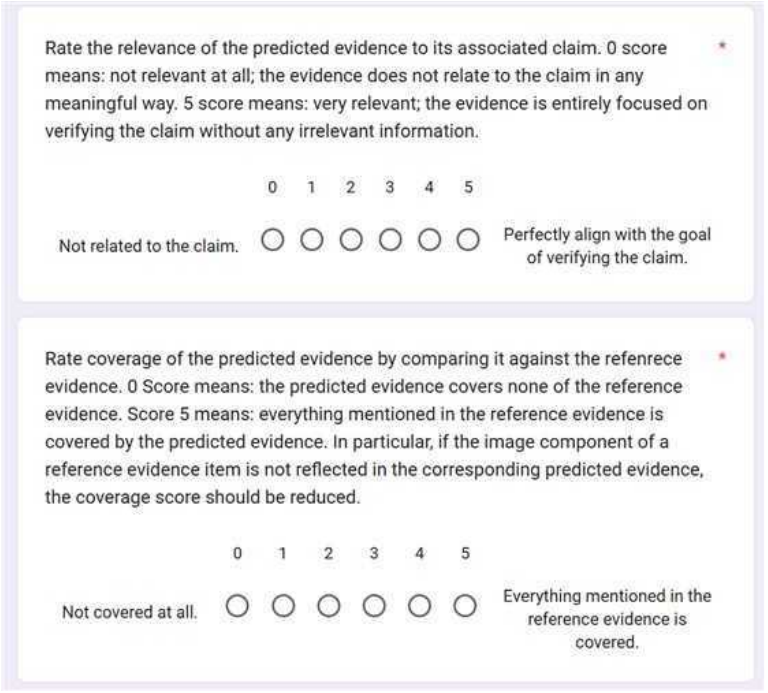}
     \caption{\textbf{The guidelines for scoring the predicted evidence on coverage and relevance.}}
     \label{fig:anno-score}
 \end{figure} 

\begin{table}[ht!]
    \centering
    \small
    \begin{tabular}{lcc}
    \toprule
     \textbf{Team} & \textbf{Avg. Coverage} & \textbf{Leaderboard \#}\\
    \midrule
     HUMANE & 4.0 & 1.\\
     ADA-AGGR & 3.5 & 3. \\
     AIC CTU& 3.84 & 2. \\
     Xxp & 2.36 & 4. \\
     REVEAL & 1.9 & 5. \\
    \bottomrule
    \end{tabular}
    \caption{\textbf{Average semantic \textbf{coverage} scores assigned to evidence samples from selected teams based on human evaluation}, next to \our \textbf{rank} the team obtained in the shared task.}
    \label{tab:human_eval_scores}
\end{table}

\section{Human Alignment Check}
\label{sec:app-human}
We conducted a human evaluation of predicted evidence by comparing it against reference (i.e., gold) evidence. As discussed in Section~\ref{sec:human-eval}, we observed limited agreement between the automatic evaluation scores and human judgments. To better understand the sources of this misalignment, we performed a fine-grained alignment analysis across evidence predicted by different teams as well as across predictions with varying automatic evaluation scores. The results of this analysis are reported in Table~\ref{tab:sepa_correlation}.

\begin{figure*}[ht!]
     \centering
     \includegraphics[scale=0.72]{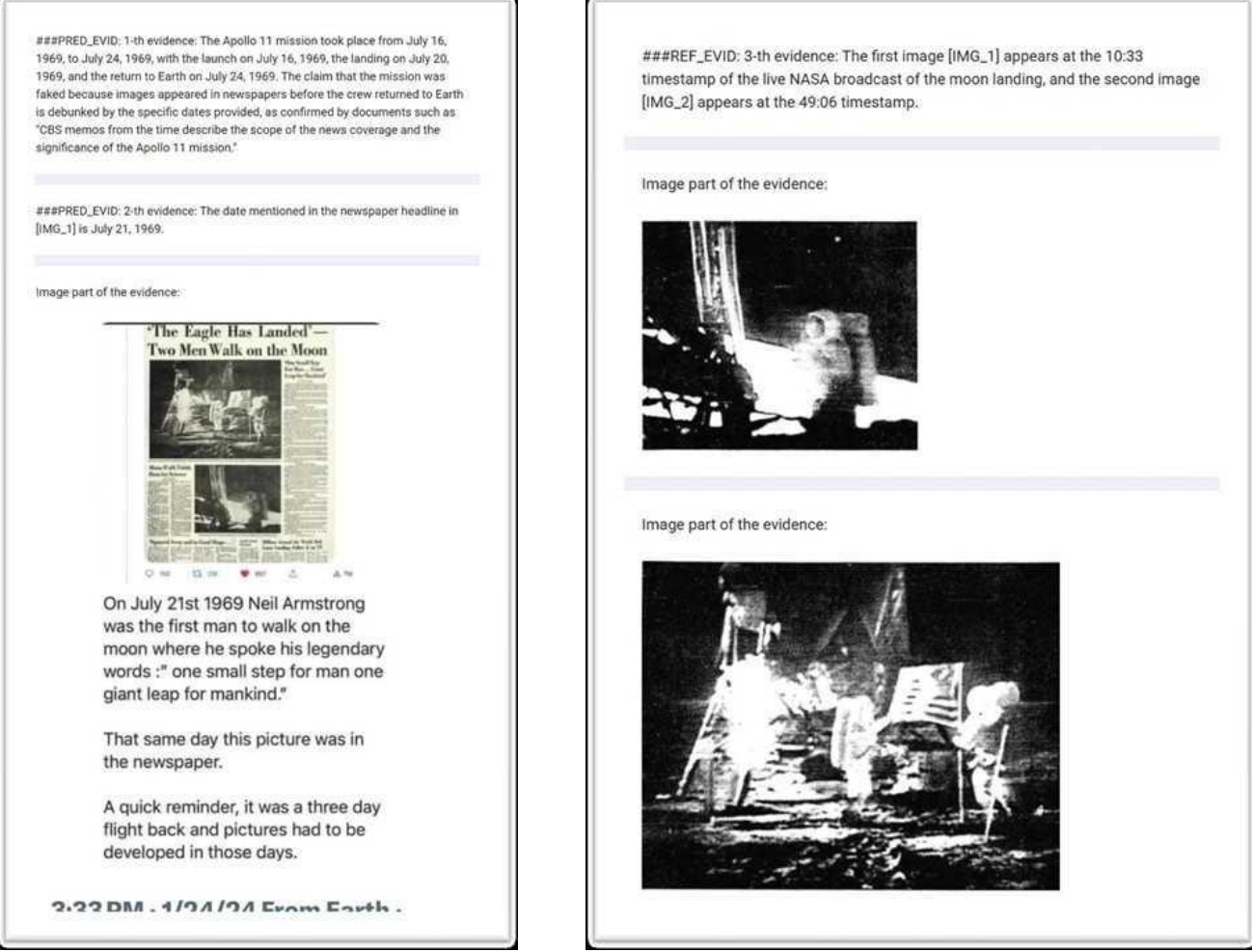}
     \caption{\textbf{An example of predicted and the corresponding reference evidence  to human annotators.} Only a subset of the evidence is shown due to length constraints.}
     \label{fig:anno-evid}
 \end{figure*}

The annotation guidelines are illustrated in Figure~\ref{fig:anno-guideline}.
Figure~\ref{fig:anno-claim} demonstrates an example of claim shown to annotators.
Figure~\ref{fig:anno-evid} shows how the predicted and reference evidence are presented to annotators.
The Figure~\ref{fig:anno-score} illustrates the two scoring dimensions, coverage and relevance, used in the human evaluation.

\begin{table*}[ht]
    \centering
    \begin{tabular}{p{0.3\linewidth} p{0.63\linewidth}} \toprule
Query type & Description \\\midrule
Generated questions & \textit{Questions are generated with gpt-3.5-turbo based on the claim. Three claim-question pairs from the training set are used as in-context examples}. \\
Generated background queries & \textit{Queries are generated with gpt-3.5-turbo based on the claim. The prompt focuses on background information, such as details about entities in the claim. Three manually constructed claim-query pairs are used as in-context examples}.\\
Generated provenance queries & \textit{Queries are generated with gpt-3.5-turbo based on the claim. The prompt focuses on information necessary to establish provenance, such as whether the claim source is a satire site. Three manually constructed claim-query pairs are used as in-context examples}. \\
Claim named entities & \textit{Named entities from the claim are extracted and used as search queries. One query for each entity is constructed, along with one query containing all entities.} \\
Most similar gold evidence & \textit{The most similar paragraph in the gold evidence document is selected using BM25, and used as a search query.} \\
Gold URL generated questions &  \textit{Queries are generated with gpt-3.5-turbo based on the URL of the gold evidence. The prompt tried to generate questions that would retrieve the URL in question. Three manually constructed URL-query pairs are used as in-context examples}.\\
Different event same entity &  \textit{Queries are generated with gpt-3.5-turbo based on the named entities in the claim. The prompt focuses on different events involving some of the same entities. Results are used as distractors to make the retrieval task harder}. \\
Similar entities & \textit{Queries are generated with gpt-3.5-turbo based on the claim. The prompt replaces entities in the claim with other similar entities, such as changing one city to another. Results are used as distractors to make the retrieval task harder}. \\
Gold questions & \textit{Gold questions used verbatim as search queries.} \\
Claim + gold question & \textit{Gold questions used verbatim as search queries. The claim is prepended, processed as in \citet{DBLP:conf/nips/SchlichtkrullG023}.} \\
Rephrased gold questions & \textit{Gold questions are rephrased using gpt-3.5-turbo, and then input as search queries.} \\ 
Gold answers & \textit{Gold questions used verbatim as search queries.} \\
Rephrased gold answers & \textit{Gold answers are rephrased using gpt-3.5-turbo, and then input as search queries.} \\\bottomrule
    \end{tabular}
    \caption{\textbf{Types of textual query input to the Google Search API for each claim in order to build the knowledge store.} Following \cite{DBLP:conf/nips/SchlichtkrullG023}, we restrict search results to documents published before the claim. For each claim, we also extend the knowledge store with the corresponding gold evidence documents.}
    \label{table:knowledge_store_prompts}
\end{table*}

\begin{table*}[ht]
    \centering
    \begin{tabular}{l|ccc|cc|cc}
    \toprule
     \textbf{Dimension}& \textbf{HUMANE}$^*$ & \textbf{HUMANE} & \textbf{REVEAL} & \textbf{Lowest} & \textbf{Highest}&\textbf{Exc.}&\textbf{All} \\
    \midrule
     $\rho$ & 0.483&0.06 &0.575 & 0.305&0.347 &0.369 &0.310\\
     $r$ &0.655 &0.04 &0.520 & 0.351& 0.370&0.392 &0.325\\
    \bottomrule
    \end{tabular}
    \caption{\textbf{Correlation with the Spearman ($\rho$) and Pearson ($r$) correlation coefficients between \our scores and human-rated scores regarding different predictions.} We reported correlation on predictions from the HUMANE and REVEAL team as well as predictions with scoring of 0 (Lowest) and 1 (Highest) from our automatic evidence evaluation model. The column HUMANE$^*$ reports correlation between human annotators. We also report correlations between human rating and our evaluation model rating when excluding (Exc.) HUMANE's predictions.
    }
    \label{tab:sepa_correlation}
\end{table*} 
Although we observed relatively low alignment between human judgments and our automatic evidence evaluation scores, human assessments of predicted evidence largely agree with the overall ranking of participating systems, as reported in Table~\ref{tab:human_eval_scores}. 
This alignment analysis suggests that while our evidence evaluation method provides a reasonable and informative baseline for assessing predicted evidence quality, it remains relatively coarse-grained and would benefit from further refinement to better capture nuanced or partially correct evidence.
\end{document}